\documentclass[runningheads]{llncs}
\usepackage[T1]{fontenc}
\usepackage{graphicx}
\usepackage{hyperref} 
\usepackage{color}

\usepackage{amsmath,amssymb,amsfonts}
\usepackage{algorithmic}

\usepackage{algorithm}
\usepackage{lineno}
\def\etal{\MakeLowercase{\textit{et~al.}}}

\newcommand{\beginsupplement}{%
	\setcounter{table}{0}
	\renewcommand{\thetable}{S\arabic{table}}%
	\setcounter{figure}{0}
	\renewcommand{\thefigure}{S\arabic{figure}}%
	\setcounter{section}{0}
	\renewcommand\thesection{\Alph{section}}%
}

\begin{document}
	
	\title{Censor-aware Semi-supervised Learning for Survival Time Prediction from Medical Images
		\thanks{This work was supported by the Australian Research Council through grants DP180103232 and FT190100525.}
	}
	
	\author{Renato Hermoza$^{\dagger}$ \qquad Gabriel Maicas$^{\dagger}$ \qquad Jacinto C. Nascimento$^{\ddagger}$ \qquad Gustavo Carneiro$^{\dagger}$}
	
	
	\institute {$^{\dagger}$Australian Institute for Machine Learning, The University of Adelaide \\ $^{\ddagger}$Institute for Systems and Robotics, Instituto Superior Tecnico, Portugal}
	
	\maketitle
	
	\begin{abstract}
		Survival time prediction from medical images is important for treatment planning, where accurate estimations can improve healthcare quality.
		One issue affecting the training of survival models is censored data.
		Most of the current survival prediction approaches are based on Cox models that can deal with censored data, but their application scope is limited because they output a hazard function instead of a survival time.
		On the other hand, methods that predict survival time usually ignore censored data, resulting in an under-utilization of the training set.
		In this work, we propose a new training method that predicts survival time using all censored and uncensored data.
		We propose to treat censored data as samples with a lower-bound time to death and estimate pseudo labels to semi-supervise a censor-aware survival time regressor.
		We evaluate our method on pathology and x-ray images from the TCGA-GM and NLST datasets.
		Our results establish the state-of-the-art survival prediction accuracy on both datasets.
		
		
	\end{abstract}

	\section{Introduction}
	\label{sec:intro}
	
	Survival time prediction models estimate the time elapsed from the start of a study until an event (e.g., death) occurs with a patient.
	These models
	are important because they may influence treatment decisions that affect the health outcomes
	of patients~\cite{cheon_accuracy_2016}.
	Thus, automated models that produce accurate survival time estimations can be beneficial to improve the quality of healthcare.
	
\begin{figure}[!t]
		\centerline{\includegraphics[width=0.90\linewidth]{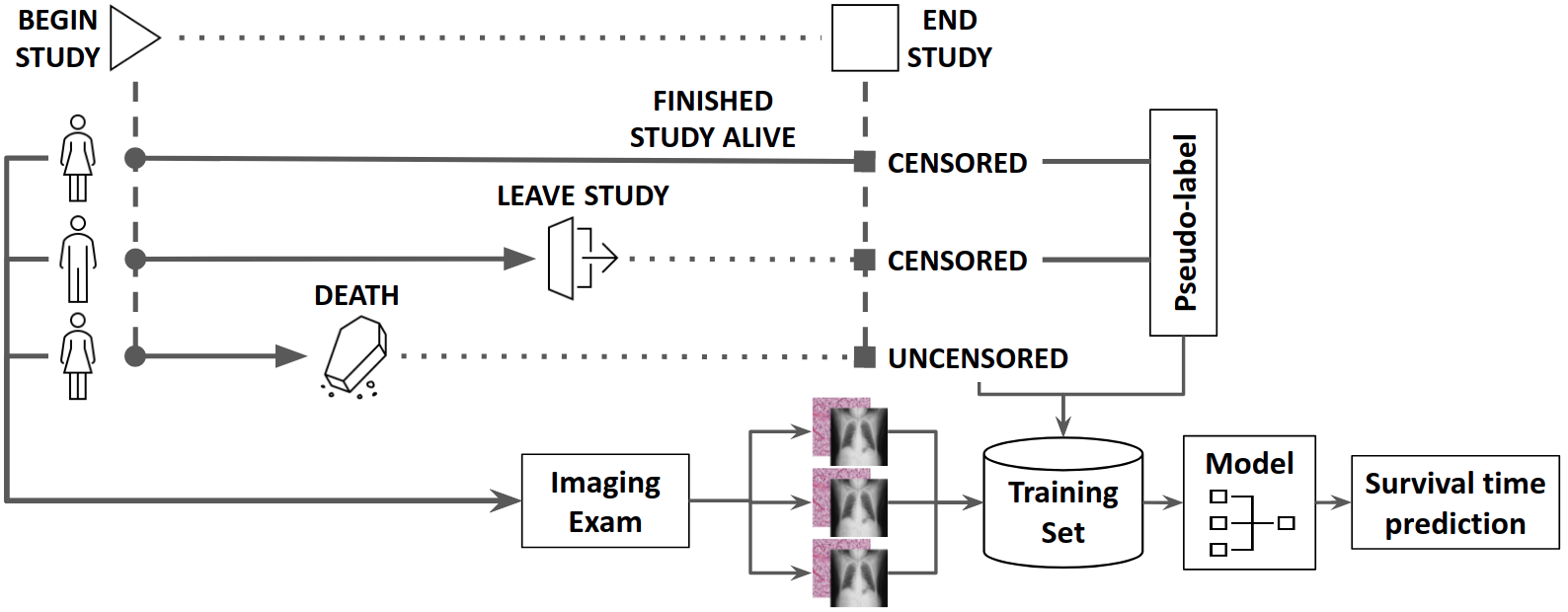}}
		\caption{At the beginning of the study, all patients are scanned for a pathology or chest x-ray image, and then patients are monitored until the end of study. Patients who die during the study represent uncensored data, while patients who do not die or leave before the end of study denote right-censored data.
			We train our model with uncensored and pseudo-labeled censored data to semi-supervise a censor-aware regressor to predict survival time.}
		\label{fig:model_intro}
	\end{figure}

	Survival prediction analysis requires the handling of right-censored samples, which represent the cases where the event of interest has not occurred either because the study finished before the event happened, or the patient left the study before its end (Fig.~\ref{fig:model_intro}).
	Current techniques to deal with censored data are typically based on Cox proportional hazards models~\cite{cox_regression_1972} that 
	rank patients in terms of their risks
	instead of predicting
	survival time, limiting their usefulness for general clinical applications~\cite{baid_overall_2020}. The problem is that the survival time prediction from a Cox hazard model requires the estimation of a baseline hazard function, which is
	is challenging to obtain in practice~\cite{xiao_censoring-aware_2020}
	A similar issue is observed in~\cite{raykar_ranking_2007} that treats survival analysis as a ranking problem without directly estimating survival times.
	
	Some methods~\cite{agravat_brain_2019,hermoza_post-hoc_2021,tang_pre-operative_2019} directly predict survival time, but they ignore the censored data for training.
	Other methods~\cite{jing_ranksurv_2019,xiao_censoring-aware_2020,wei1992accelerated} use censored data during training, but in a sub-optimally manner.
	More specifically, these models use the censoring time
	as a lower bound for the event of interest. 
	Even though this is better than disregarding the censored cases,
	such approaches do not consider the potential differences between the censoring time and the hidden survival time.
	We argue that if the values of these differences can be estimated with pseudo-labeling mechanisms, survival prediction models could be more accurate.
	However, pseudo labels estimated for the censored data
	may introduce noisy labels for training, which have been studied for classification~\cite{li2020dividemix,liu_early-learning_2020} and segmentation~\cite{zheng2021rectifying}, but never for survival prediction in a semi-supervision context.

	In this paper, we propose a new training method for deep learning models to predict survival time from medical images using all censored and uncensored training data.
	The main contributions of this paper are:
	\begin{itemize}
		\item a method that estimates a pseudo label for the survival time of censored data (lower bounded by the annotated censoring time) that is used to semi-supervise a censor-aware survival time regressor; and
		\item two new regularization losses to handle pseudo label noise: a) we adapt the early-learning regularization (ELR)~\cite{liu_early-learning_2020} loss from classification to survival prediction; and b) inspired by risk prediction models, we use a censor-aware ranking loss to produce a correct sorting of samples in terms of survival time.
	\end{itemize}
	We evaluate our method using the TCGA-GM~\cite{mobadersany_predicting_2018} dataset of pathological images, and the NLST dataset~\cite{national2011reduced,national2011national} of chest x-ray images.
	Our results show a clear benefit of using pseudo labels to train survival prediction models, achieving the new state-of-the-art (SOTA) survival time prediction results for both datasets.
	We make our code available at \url{https://github.com/renato145/CASurv}.
	\section{Method}
	\label{sec:method}
	
	\begin{figure}[!t]
		\centerline{\includegraphics[width=0.9\linewidth]{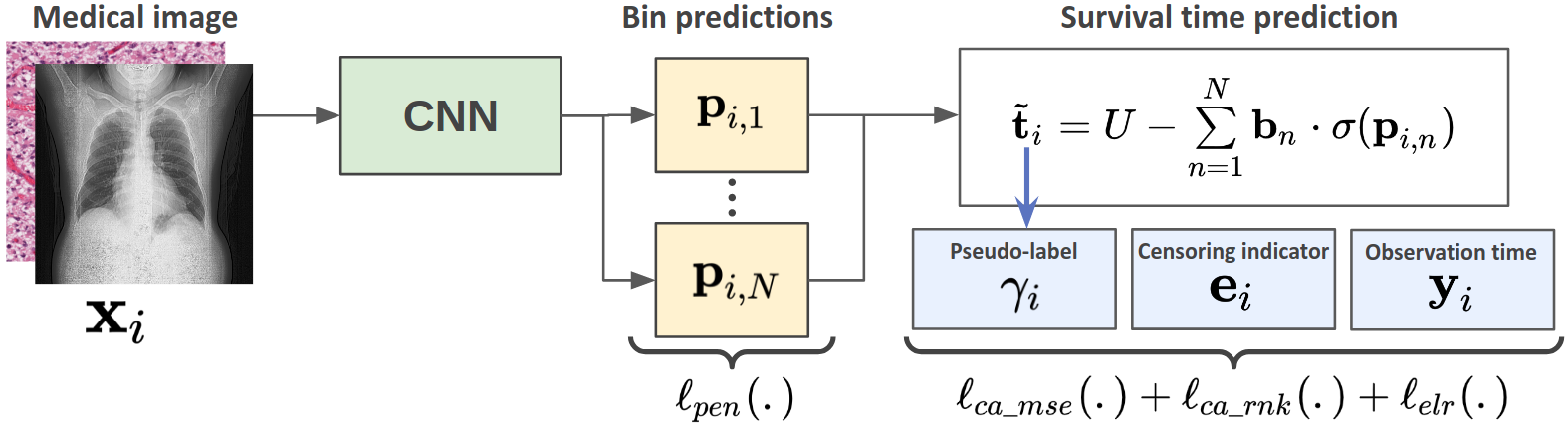}}
		\caption{
			The proposed model outputs a set of $N$ bin predictions $\{\mathbf{p}_{i,n}\}_{n=1}^{N}$ (each bin representing an amount of time $\mathbf{b}_n$) that are aggregated to produce a survival time prediction for the $i^{th}$ case.
			This prediction is achieved by taking the maximum survival time $U$ and subtracting it by the activation of each bin $\sigma(p_{i,n})$ times the amount of time in $\mathbf{b}_n$.
			A set of loss functions are used while training: a censored-aware version of MSE ($\ell_{ca\_mse}$), a penalization term for bin consistency ($\ell_{pen}$), a rank loss ($\ell_{ca\_rnk}$), and an adapted version of the ELR regularization~\cite{liu_early-learning_2020} to survival prediction ($\ell_{elr}$).
		}
		\label{fig:model}
	\end{figure}

	To explain our method, we define the data set as
	$\mathcal{D} = \{(\mathbf{x}_i,\mathbf{y}_i,\mathbf{e}_i)\}_{i=1}^{|\mathcal{D}|}$,
	where
	$\mathbf{x}_i \in \mathcal{X}$
	denotes a medical image with $\mathcal{X} \subset \mathbb{R}^{H \times W \times C}$ ($H,W,C$ denote image height, width and number of color channels),
	$\mathbf{y}_i \in \mathbb N$ indicates the observation time defined in days, and $\mathbf{e}_i \in \{0,1\}$ is the censoring indicator.
	When $\mathbf{e}_i = 0$ (uncensored observation), $\mathbf{y}_i$ corresponds to a survival time $\mathbf{t}_i$ which indicates the event of death for the individual.
	In cases where $\mathbf{e}_i = 1$ (censored observation) $\mathbf{t}_i$ is unknown, but is lower bounded by $\mathbf{y}_i$ (i.e, $\mathbf{t}_i > \mathbf{y}_i$).
	
	\subsection{Model}
	\label{sec:model}

	The architecture of our model extends the implementation from~\cite{hermoza_post-hoc_2021} to work with censored data (Fig.~\ref{fig:model}). Our survival time prediction model uses a Convolutional Neural Network (CNN)~\cite{lecun_gradient-based_1998} represented by
	\begin{equation}
		f:\mathcal{X} \times \theta \to \mathcal{P},
		\label{eq:inference}
	\end{equation}
	which is parameterized by $\theta \in \Theta$ and takes an image $\mathbf{x}$ to output a survival time confidence vector $\mathbf{p} \in \mathcal{P}$ that represents the confidence on regressing to the number of days in each of the $N$ bins of $\mathcal{P} \subset \mathbb{R}^N$ that discretize $\mathbf{y}_i$.
	Each bin of the vector $\mathbf{b} \in\mathbb{N}^N$ represents a number of days interval and the sum of them is the upper limit for the survival time in the dataset: $U = \sum^{N}_{n=1} {\mathbf{b}(n)}$.
	Bins are discretized non-uniformly to balance number of samples per bin to avoid performance degradation~\cite{xiao_censoring-aware_2020}.

	To obtain the survival time prediction we first calculate the survival number of days per bin $n$, as in $\tilde{\mathbf{p}}({n}) = \mathbf{b}(n) \cdot \sigma(\mathbf{p}({n}))$,
	where $\sigma(.)$ represents the sigmoid function. The survival time prediction is obtained with $\tilde{\mathbf{t}}= h(\mathbf{x};\theta) = U - \sum\limits^{N}_{n=1} \tilde{\mathbf{p}}({n})$,
	where $h(\mathbf{x};\theta)$ represents the full survival time regression model.
	Hence, a larger activation $\mathbf{p}({n})$ indicates higher risk, as the predicted $\tilde{\mathbf{t}}$ will be smaller.
	The training of our model minimizes the loss function:
	\begin{equation}
		\scalebox{0.96}{
			$\begin{split}
			\ell(\mathcal{D},\theta) =
			\frac{1}{|\mathcal{D}|} \sum_{(\mathbf{x}_i,\mathbf{y}_i,\mathbf{e}_i) \in \mathcal{D}} &[
			\ell_{ca\_mse}(\tilde{\mathbf{t}}_i,\mathbf{y}_i,\mathbf{e}_i,\mathbf{s}_i)
			+ \alpha \cdot \ell_{elr}(\mathbf{p}_i)
			+ \beta \cdot \ell_{pen}(\mathbf{p}_i) \\
			& + \frac{1}{|\mathcal{D}|}\sum_{j=1}^{|\mathcal{D}|}\delta \cdot \ell_{ca\_rnk}(\tilde{\mathbf{t}}_i,\tilde{\mathbf{t}}_j,\mathbf{y}_i,\mathbf{y}_j,\mathbf{e}_i,\mathbf{e}_j)]
			\end{split}$
		}
		\label{eq:finalLoss}
	\end{equation}
	where $\ell_{ca\_mse}(.)$ is a censor-aware mean squared error defined in~\eqref{eq:camae}, $\mathbf{s}_i$ indicates if sample $i$ is pseudo-labeled, $\ell_{elr}(.)$ is the ELR regularization~\cite{liu_early-learning_2020} adapted for survival time prediction defined in~\eqref{eq:elr},
	$\ell_{pen}(.)$ is the penalization term defined in~\eqref{eq:pen_loss},
	$\ell_{ca\_rnk}(.)$ is a censor-aware rank loss defined in~\eqref{eq:rank}, and
	$\alpha$, $\beta$ and $\delta$ control the strength of $\ell_{elr}(.)$, $\ell_{pen}(.)$ and
	$\ell_{ca\_rnk}(.)$ losses.

	\subsubsection{Pseudo Labels}
	We proposed the use of pseudo labels to semi-supervise our censor-aware survival time regressor.
	For a censored observation $i$, a pseudo label $\gamma_i$ is estimated as:
	\begin{equation}
		\gamma_i = \max(\mathbf{y}_i, \tilde{\mathbf{t}}_i),
		\label{eq:pseudolbl}
	\end{equation}
	which takes advantage of the nature of censored data, as being lower bounded by $\mathbf{y}_i$. We estimate pseudo labels at the beginning of each epoch and treat them as uncensored observations by re-labeling $\mathbf{y}_i = \gamma_i$ and setting $\mathbf{s}_i=1$.

	The quality of generated pseudo labels depends on the training procedure stage, where pseudo labels produced during the first epochs are less accurate than the ones at the last epochs.
	Hence, we control the ratio of censored sample labels to be replaced with pseudo labels at each epoch using a cosine annealing schedule to have no pseudo labels at the beginning of the training and all censored data with their pseudo labels by the end of the training.

	\subsubsection{Censor-Aware Mean Squared Error (CA-MSE)}
	Using a regular MSE loss is not suitable for survival prediction because 
	the survival time for censored observations is unknown, but lower-bounded by $\mathbf{y}_i$.
	Therefore, we assume an error exists for censored samples when $\tilde{\mathbf{t}}_i < \mathbf{y}_i$.
	However, when $\tilde{\mathbf{t}}_i > \mathbf{y}_i$, it will be incorrect to assume an error, as the real survival time is unknown.
	To mitigate this issue, we introduce the CA-MSE loss:
	\begin{equation}
		\scalebox{0.89}{
			$\ell_{ca\_mse}(\tilde{\mathbf{t}}_i,\mathbf{y}_i,\mathbf{e}_i,\mathbf{s}_i) = \left\{\begin{array}{cl}
			0, & \text { if } (\mathbf{e}_i=1) \land (\tilde{\mathbf{t}}_i>\mathbf{y}_i) \\
			\tau^{\mathbf{s}_i}(\tilde{\mathbf{t}}_i - \mathbf{y}_i)^2, & \text { otherwise}
			\end{array}\right.,$
		}
		\label{eq:camae}
	\end{equation}
	where $\tau \in (0,1)$
	reduces the weight for pseudo-labeled data ($\mathbf{s}_i=1$).
	
	
	\subsubsection{Early-Learning Regularization (ELR)}
	To deal with noisy pseudo labels, we modify ELR~\cite{liu_early-learning_2020} to work in our survival time prediction setting, as follows:
	\begin{equation}
		\ell_{elr}(\mathbf{p}_i) = \log \left (1 - (1/N) \left ( \sigma(\mathbf{p}_i)^{\top}\sigma(\mathbf{q}_i) \right ) \right ),
		\label{eq:elr}
	\end{equation}
	where $\mathbf{q}_i^{(k)} = \psi\mathbf{q}_i^{(k-1)} + (1-\psi)\mathbf{p}_i^{(k)}$ (with $\mathbf{q}_i \in \mathcal{P}$) is the temporal ensembling momentum~\cite{liu_early-learning_2020} of the survival predictions, with $k$ denoting the training epoch, and  $\psi\in[0,1]$.
	The idea of the loss in~\eqref{eq:elr} is to never stop training for samples where the model prediction coincides with the temporal ensembling momentum (i.e., the 'clean' pseudo-labeled samples),
	and to never train for the noisy pseudo-labeled samples~\cite{liu_early-learning_2020}.

	\subsubsection{Bin Penalization Term}
	We add the penalization term described in~\cite{hermoza_post-hoc_2021} that forces a bin $n$ to be active only when all previous bins ($1$ to $n-1$) are active, forcing bins to represent sequential risk levels:
	\begin{equation}
		\ell_{pen}(\mathbf{p}_i) = \frac{1}{N-1} \sum\limits^{N-1}_{n=1} \max(0, (\sigma(\mathbf{p}_{i}(n+1)) - \sigma(\mathbf{p}_{i}(n)))).
		\label{eq:pen_loss}
	\end{equation}

	\subsubsection{Censor-Aware Rank Loss}
	Cox proportional hazards models~\cite{cox_regression_1972} aim to rank training samples instead of the directly estimating survival time.
	Following this reasoning, we propose a censor-aware ranking loss to encourage this sample sorting behavior.
	The rank loss has the following form:
	\begin{equation}
		\ell_{ca\_rnk}(\tilde{\mathbf{t}}_i,\tilde{\mathbf{t}}_j,\mathbf{y}_i,\mathbf{y}_j,\mathbf{e}_i,\mathbf{e}_j) = \max(0, -\mathcal{G}(\mathbf{y}_i,\mathbf{y}_j,\mathbf{e}_i,\mathbf{e}_j) \times (\tilde{\mathbf{t}}_i - \tilde{\mathbf{t}}_j))
		\label{eq:rank}
	\end{equation}
	where $i$ and $j$ index all pairs of samples in a training mini-batch, and
	\begin{equation}
		\begin{split}
			\mathcal{G}(\mathbf{y}_i,\mathbf{y}_j,\mathbf{e}_i,\mathbf{e}_j) = \left\{\begin{array}{cl}
				+1, & \text { if } (\mathbf{y}_i > \mathbf{y}_j) \land \mathbf{e}_j=0 \\
				-1, & \text { if } (\mathbf{y}_i \le \mathbf{y}_j) \land \mathbf{e}_i=0 \\
				0, & \text { otherwise}
			\end{array}\right..
		\end{split}
		\label{eq:rank_target}
	\end{equation}
	In~\eqref{eq:rank_target}, when $\mathcal{G}(.) = +1$ the sample $i$ should be ranked higher than sample $j$, when $\mathcal{G}(.) = -1$ the sample $j$ should be ranked higher than sample $i$, and the loss is ignored when $\mathcal{G}(.) = 0$.

	\section{Experiments}
	\label{sec:exp}
	
	In this section, we describe the datasets and explain the experimental setup and evaluation measures, followed by the presentation of results.
	
	\subsection{Data Sets}
	
	\subsubsection{TCGA-GM dataset}
	
	
	The Cancer Genome Atlas (TCGA) Lower-Grade Glioma (LGG) and Glioblastoma (GBM) projects: TCGA-GM~\cite{mobadersany_predicting_2018} consists of 1,505 patches extracted from 1,061 whole-slide pathological images.
	A total of 769 unique patients with gliomas were included in the study with 381 censored and 388 uncensored observations.
	The labels are given in days and can last from 1 to 6423 days.
	We use the published train-test split provided with the data set~\cite{mobadersany_predicting_2018}.

	\subsubsection{NLST dataset}
	

	The National Lung Screening Trial (NLST)~\cite{national2011reduced,national2011national} is a randomized multicenter study 
	for early detection of lung cancer of current or former heavy smokers of ages 55 to 74.
	We only used the chest x-ray images from the dataset and excluded cases where the patient died for causes other than lung cancer.
	The original study includes 25,681 patients (77,040 images), and after filtering, we had 15,244 patients (47,947 images) with 272 uncensored cases.
	The labels are given in days and can last from 1 to 2983 days.
	We patient-wise split the dataset into training (9,133 patients), validation (3,058 patients) and testing (3,053 patients) sets, maintaining the same demographic distribution across sets.

	\subsection{Experimental Set Up}
	
	The input image size is adjusted to $512^2$ pixels for both datasets and normalized with ImageNet~\cite{russakovsky_imagenet_2015} mean and standard deviation.
	For the model in~\eqref{eq:inference}, the space $\mathcal{P}$ has 5 bins for TCGA-GM pathology images and 3 bins for NLST chest x-ray images.
	The model $f(\mathbf{x};\theta)$ in~\eqref{eq:inference} is an ImageNet pre-trained ResNet-18~\cite{he_deep_2016}.
	The training of the model uses Adam~\cite{kingma_adam:_2015} optimizer with a momentum of 0.9, weight decay of 0.01 and a mini-batch size of 32 for 40 epochs.
	To obtain pseudo labels we have $\tau=0.5$ in~\eqref{eq:camae}, the bin penalization term $\beta=1e6$ in~\eqref{eq:pen_loss}, the ELR~\cite{liu_early-learning_2020} loss in~\eqref{eq:elr} with $\alpha=100$ and $\psi=0.5$, and the rank loss in~\eqref{eq:rank} with $\delta=1$.
	We run all experiments using PyTorch~\cite{paszke_pytorch_2019} v1.8 on a NVIDIA V100 GPU.
	The total training time is 45 minutes for the TCGA-GM dataset and 240 minutes for the NLST dataset, and the inference time for a single x-ray image takes 2.8 ms.
	
	To evaluate our method, we use the Mean Absolute Error (MAE) between our predictions and the observed time, where the error is reported patient-wise, aggregating the mean of all image predictions per patient.
	To account for censored data, we use the following metric proposed by Xiao \etal~\cite{xiao_censoring-aware_2020}.
	\begin{equation}
		\text{MAE}=\mathbb{E}\left [ |\tilde{\mathbf{t}}-\mathbf{y}| \mathbb{I}({\tilde{\mathbf{t}}<\mathbf{y}})+(1-\mathbf{e})|\tilde{\mathbf{t}}-\mathbf{y}|\mathbb{I}({\tilde{\mathbf{t}} \geq \mathbf{y}}) \right ],
		\label{eq:mae}
	\end{equation}
	where $\mathbb{I}$ is the indicator function. In~\eqref{eq:mae}, the censored cases where the model predicts the survival time $\tilde{\mathbf{t}} \ge \mathbf{y}$ are not counted as errors.
	As a result, a higher count of censored records will decrease the average MAE, so we also measure the MAE considering only uncensored records.
	We also report concordance index (C-index)~\cite{harrell_evaluating_1982} to measure the correct ranking of the predictions. 

	\subsection{Results}
	\label{sec:exp_results}
	
	In Table~\ref{tab:res_tcga_nlst}, we compare survival time prediction results between our method and SOTA methods~\cite{mobadersany_predicting_2018,xiao_censoring-aware_2020,zhu_deep_2016} on TCGA-GM, and DeepConvSurv~\cite{zhu_deep_2016} on NLST.
	Our method sets the new SOTA results in terms of MAE (all samples and only uncensored ones) and C-index for both datasets. The low MAE result for our approach on NLST can be explained by the fact that the vast majority of cases are censored (over 95\%), and our method successfully estimates a survival time larger than the censoring time, producing $\text{MAE}=0$ for many censored cases, according to~\eqref{eq:mae}. 
	
	\begin{table}[t!]
		\centering
		\caption{
			Comparison of survival time prediction between our approach and SOTA methods on the TCGA-GM and NLST. Evaluation metrics are mean absolute error (MAE) in days, and concordance index (C-index). Best results are highlighted.
		}
		\label{tab:res_tcga_nlst}
		\setlength{\tabcolsep}{3pt}
		\scalebox{0.82}{
			\begin{tabular}{l|c|c|c}
				\hline
				Method & ~MAE (all samples)~ & ~MAE (only uncensored)~ & ~C-index~ \\
				\hline
				\multicolumn{4}{c}{TCGA-GM} \\
				\hline
				DeepConvSurv~\cite{zhu_deep_2016} & ~439.1~ & - & ~0.731~ \\
				AFT~\cite{wei1992accelerated} & ~386.5~ & - & ~0.685~ \\
				SCNNs~\cite{mobadersany_predicting_2018} & ~424.5~ & - & ~0.725~ \\
				CDOR~\cite{xiao_censoring-aware_2020} & ~321.2~ & - & ~0.737~ \\
				Ours & \textbf{286.95} & \textbf{365.06} & \textbf{0.740} \\
				\hline
				\multicolumn{4}{c}{NLST} \\
				\hline
				DeepConvSurv~\cite{zhu_deep_2016} & 27.98 & 1334.22 &   0.690 \\
				Ours  &    \textbf{26.28} & \textbf{1275.24} & \textbf{0.756} \\
				\hline
			\end{tabular}
		}
	\end{table}

	On Table~\ref{tab:ablations}, we show the ablation study for the components explained in Section~\ref{sec:model}, where we start from a baseline model Base trained with $\ell_{ca\_mse}$ and $\ell_{pen}$. Then we show how results change with the use of $\ell_{ca\_rnk}$ (RankLoss), pseudo labels and $\ell_{elr}$ (ELR).
	Regarding the MAE results on TCGA-GM and NLST datasets, we can observe a clear  benefit of using pseudo labels.
	For TCGA-GM, the use of pseudo labels improve MAE by more than 20\% compared with the Base model with RankLoss.
	For NLST, the smaller MAE reduction with pseudo labels can be explained by the large proportion of censored cases, which makes the robust estimation of survival time a challenging task.
	Considering the C-index, pseudo labels show similar benefits for TCGA-GM, with a slightly better result.
	On NLST, pseudo labels improve the C-index from 0.73 to 0.75 over the Base model with RankLoss.
	On both datasets, the use of
	Base model with RankLoss, pseudo labels, and ELR,
	presents either the best results or a competitive result with best result.

	\begin{table}[t!]
		\centering
		\caption{
			Ablation study for our method on the TCGA-GM and NLST datasets. The evaluation metrics are: mean absolute error (MAE) in days and concordance index (C-index). Best results are highlighted in bold.
		}
		\label{tab:ablations}
		\scalebox{0.80}{
			\begin{tabular}{c|c|c|c|c|c|c|c|c}
				\hline
				~$\ell_{ca\_mse}$~ & ~$\ell_{ca\_rnk}$~ & ~Pseudo~ & ~$\ell_{elr}$~  & MAE & Best MAE & Best MAE & C-index & Best \\
				$+\ell_{pen}$ & & labels & & (all samples) & (all samples) & (only uncensored) & & C-index \\
				\hline
				\multicolumn{9}{c}{TCGA-GM} \\
				\hline
				\checkmark & - & - & - &  379.40 ±20.67 &    356.74 & 521.91 &  0.702 ±0.009 &         0.707 \\
				\checkmark & \checkmark & - & - &  380.49 ±17.55 &    361.76 & 517.41 &  0.722 ±0.008 &         0.730 \\
				\checkmark & - & \checkmark & - &   308.01 ±7.99 &    302.04 & 405.48 &  0.725 ±0.014 &         \textbf{0.740} \\
				\checkmark & - & \checkmark & \checkmark &   311.76 ±6.54 &    304.42 & 404.32 &  0.711 ±0.016 &         0.723 \\
				\checkmark & \checkmark & \checkmark & - &   298.56 ±6.31 &    291.53 & 411.79 &  0.718 ±0.016 &         0.736 \\
				\checkmark & \checkmark & \checkmark & \checkmark &   \textbf{294.45 ±6.49} &    \textbf{286.95} & \textbf{365.06}  &  \textbf{0.728 ±0.010} &         \textbf{0.740} \\
				\hline
				\multicolumn{9}{c}{NLST} \\
				\hline
				\checkmark & - & - & - &    27.24 ±0.15 &     27.10 & 1283.41 &  0.729 ±0.004 &         0.732 \\
				\checkmark & \checkmark & - & - &    27.68 ±0.33 &     27.31 & 1286.27 &  0.732 ±0.003 &         0.736 \\
				\checkmark & - & \checkmark & - &    \textbf{26.97 ±0.97} &     \textbf{26.28} & \textbf{1275.24} &  0.751 ±0.006 &         \textbf{0.756} \\
				\checkmark & - & \checkmark & \checkmark &    29.05 ±0.59 &     28.37 & 1397.44 &  0.703 ±0.002 &         0.705 \\
				\checkmark & \checkmark & \checkmark & - &    27.83 ±2.32 &     26.29 & 1283.14 &  0.747 ±0.002 &         0.749 \\
				\checkmark & \checkmark & \checkmark & \checkmark &    27.13 ±0.86 &     26.61 & 1277.02 &  \textbf{0.752 ±0.002} &         0.754 \\
				\hline
			\end{tabular}
		}
	\end{table}

\begin{figure}[!t]
		\centerline{\includegraphics[width=1.0\linewidth]{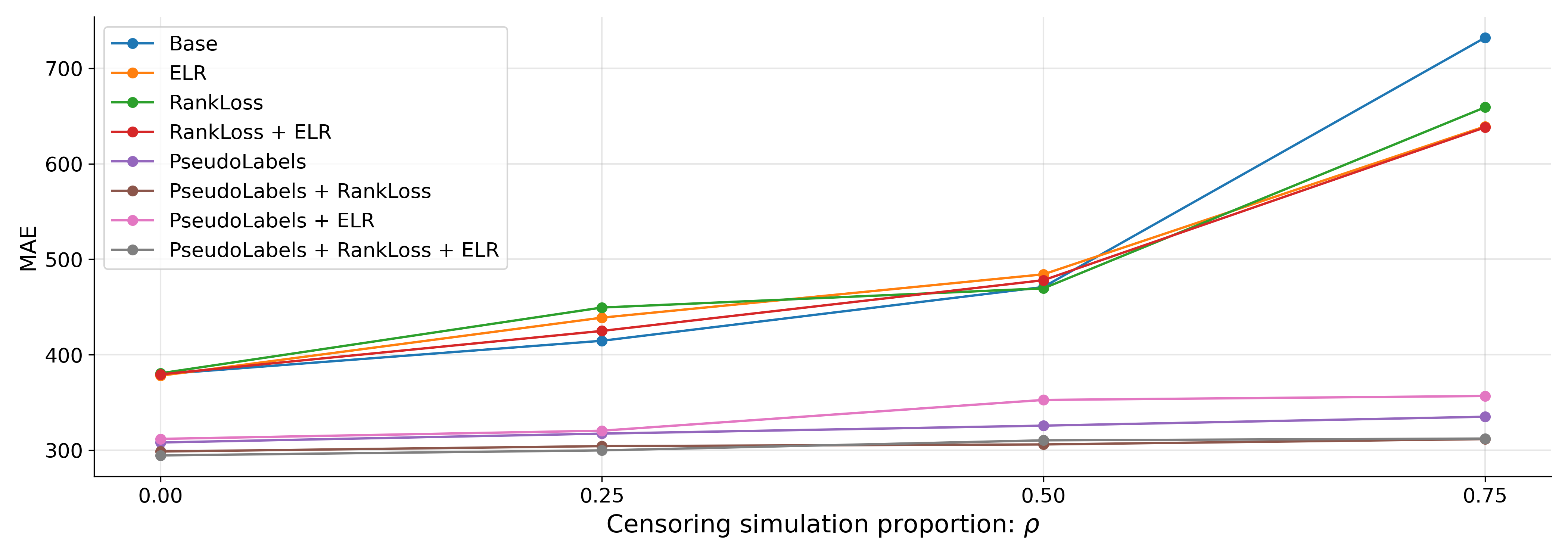}}
		\caption{
			Performance of our method under different proportions of censored vs uncensored labels on the TCGA-GM dataset.
			The x-axis shows the ratio of uncensored records transformed to be censored ($\rho$), and the y-axis shows MAE.
			Note that pseudo labels combined with RankLoss and ELR show robust results in scenarios with small percentages of uncensored records.
		}
		
		\label{fig:simulate_censoring}
	\end{figure}

	The differences between the results on NLST and TCGA-GM can be attributed to different proportions of censored/uncensored cases in both datasets.
	We test the hypothesis that such differences have a strong impact in the performance of our algorithm by formulating an experiment using the TCGA-GM dataset, where we simulate different proportion of censored records.
	Figure~\ref{fig:simulate_censoring} shows the result of this simulation, where we show MAE as a function of the proportion of censored data, denoted by $\rho$.
	The results show that the PseudoLabels methods (combined with RankLoss and ELR losses) keep the MAE roughly constant, while Base, ELR and RankLoss results deteriorate significantly. The best results are achieved by PseudoLabels + RankLoss and PseudoLabels + RankLoss + ELR, suggesting that pseudo labels and both losses are important for the method to be robust to large proportions of censored data.

	\section{Discussion and Conclusion}
	\label{sec:disc}
	
	
	Current techniques for survival prediction discard or sub-optimally use the variability present in censored data.
	In this paper we proposed a method to exploit such variability on censored data by using pseudo labels to semi-supervise the learning process.
	Experimental results in Table~\ref{tab:res_tcga_nlst} showed that our proposed method obtains SOTA results for survival time prediction on the TCGA-GM and NLST datasets.
	In fact, the effect of pseudo labels can be observed in Figure~\ref{fig:simulate_censoring}, where we artificially censored samples from the TCGA-GM dataset.
	It is clear that the use of pseudo labels provide solid robustness to a varying proportion of censored data in the training set.
	Combining the pseudo labels with a regularisation loss that accounts for noisy pseudo labels and censor-aware rank loss improves this robustness and results on datasets.
	However, is important to note that the method may not be able to produce good quality pseudo-labels on cases where the dataset contains few uncensored records.
	We expect our newly proposed method to foster the development of new survival time prediction models that exploit the variability present in censored data.

	\bibliographystyle{splncs04}
	\bibliography{zotero}
	
	\newpage
	
	\beginsupplement
	\section{Supplementary Material}
	
	\begin{algorithm}
		\caption{Pseudocode for the training procedure}
		\begin{algorithmic}
			\REQUIRE $\mathcal{D} = \{(\mathbf{x}_i,\mathbf{y}_i,\mathbf{e}_i\}_{i=1}^{|\mathcal{D}|} = \text{training data}$
			\REQUIRE $h(\mathbf{x};\theta) = \text{survival model with parameters } \theta$
			\REQUIRE $\alpha = \text{ELR loss strength}$
			\REQUIRE $\beta = \text{Bin penalization strength}$
			\REQUIRE $\delta = \text{Rank loss strength}$
			\FOR{$k \text{ in } [1,k_{total}]$}
			\STATE $\textit{// Ratio of pseudo-labels from censored samples}$
			\STATE $m \leftarrow (1 + \cos{\pi} (1-k/k_{total})) / 2$
			\STATE $\textit{// Set of censored samples to use pseudo-labels}$
			\STATE $P \leftarrow random\_sample(\mathcal{D},m)$
			\FOR{$i \text{ in } P$} 
			\STATE $\textit{// Re-label censored sample}$
			\STATE $\tilde{\mathbf{t}}_i \leftarrow h(\mathbf{x}_i;\theta)$
			\STATE $\mathbf{y}_i \leftarrow \max(\mathbf{y}_i, \tilde{\mathbf{t}}_i)$
			\STATE $\mathbf{e}_i \leftarrow 1$
			\ENDFOR
			\STATE $\textit{// Update network parameters}$
			\FOR{\text{each minibatch} B}
			\FOR{$j \text{ in } B$}
			\STATE $\tilde{\mathbf{t}}_j \leftarrow h(\mathbf{x}_j;\theta)$
			\ENDFOR
			\STATE $loss \leftarrow \frac{1}{|\mathcal{D}|} \sum_{i=1}^{|\mathcal{D}|} \ell_{ca\_mse}(\tilde{\mathbf{t}}_i,\mathbf{y}_i,\mathbf{e}_i)$ \\
			\STATE \hspace{30pt} $+\alpha \cdot \ell_{elr}(\mathbf{p}_i) + \beta \cdot \ell_{pen}(\mathbf{p}_i)$
			\STATE \hspace{30pt} $+\frac{1}{|\mathcal{D}|}\sum_{j=1}^{|\mathcal{D}|}\delta \cdot \ell_{ca\_rnk}(\tilde{\mathbf{t}}_i,\tilde{\mathbf{t}}_j,\mathbf{y}_i,\mathbf{y}_j,\mathbf{e}_i,\mathbf{e}_j)$
			\STATE $\text{update } \theta \text{ using SGD}$ with $loss$
			\ENDFOR
			\ENDFOR
			\RETURN $\theta$
		\end{algorithmic}
		\label{alg:train}
	\end{algorithm}
	
	\begin{figure}[!t]
		\centerline{\includegraphics[width=0.9\linewidth]{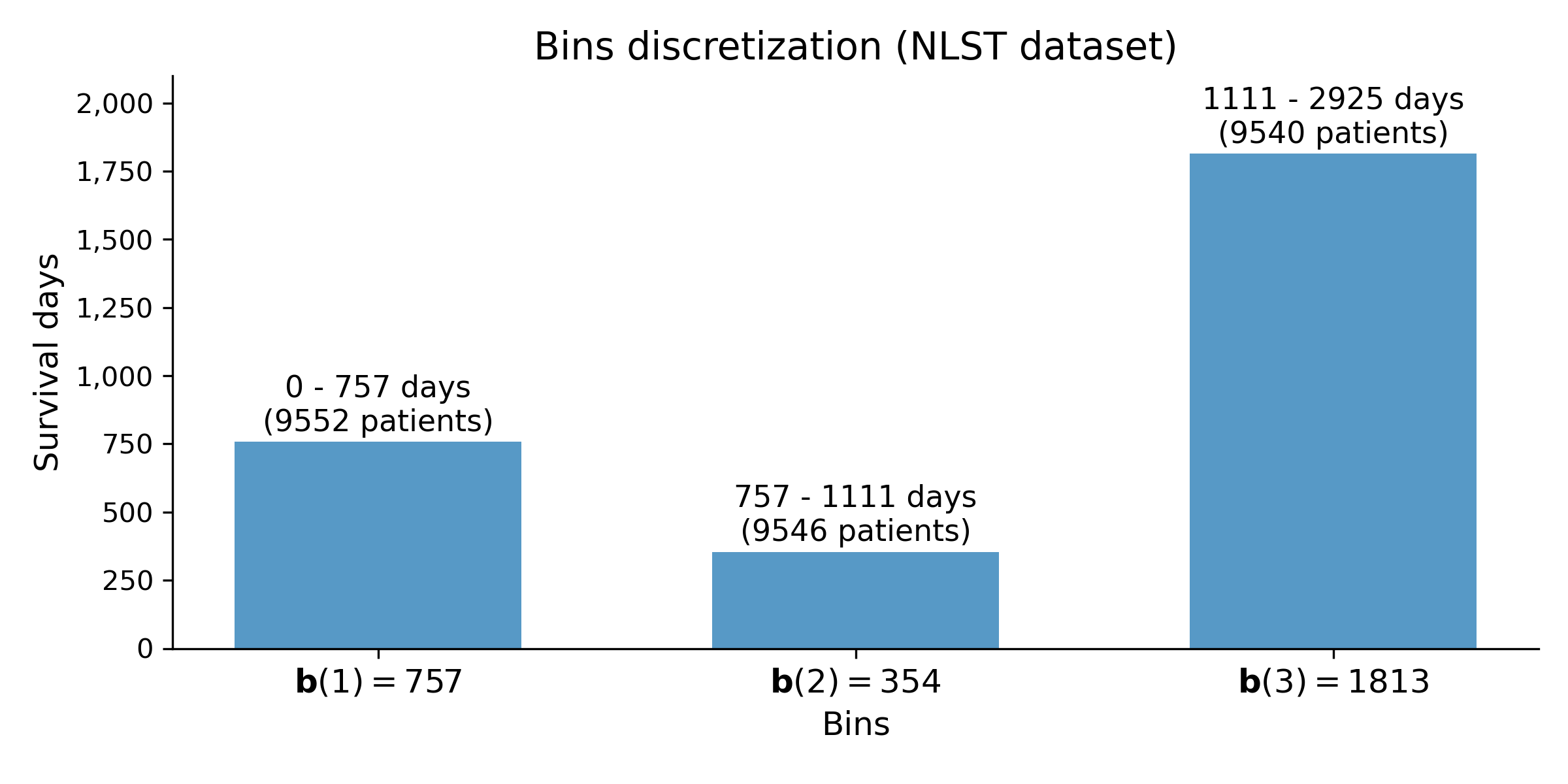}}
		\caption{
			We discretize the NLST dataset survival time into three bins, each representing a different amount of time, as depicted in the plot (e.g., $\mathbf{b}(1)=757$ days, $\mathbf{b}(2)=354$ days, and $\mathbf{b}(3)=1813$ days).
			The discretization strategy aims to balance the number of patients per bin.
			Note that this discretization strategy uses only the information in the training set.
		}
		\label{fig:bins}
	\end{figure}
	
	\begin{table}
		\centering
		\caption{
			One-sided Wilcoxon signed-rank test for the TCGA-GM dataset.
			We compare the mean absolute error (MAE) of our best performing method ($\text{Pseudolabels} + \ell_{ca\_rnk} + \ell_{elr}$) against the rest. \textsuperscript{*} indicates that the best performing method is better assuming a significance level of 5\%.
		}
		\label{tab:test_tcga}
		\scalebox{1.00}{
			\begin{tabular}{c|c|c|c|c|l}
				\hline
				~$\ell_{pen}$~ & ~$\ell_{ca\_rnk}$~ & ~Pseudo-labels~ & ~$\ell_{elr}$~  & ~Wilcox~ & ~p-value~ \\
				\hline
				\checkmark & - & - & - &  2157.0 &   ~0.0115 \textsuperscript{*} \\
				\checkmark & \checkmark & - & - &  2069.0 &   ~0.0113 \textsuperscript{*} \\
				\checkmark & - & \checkmark & - &  2383.0 &   ~0.0579 \\
				\checkmark & - & \checkmark & \checkmark &  2633.0 &   ~0.3164 \\
				\checkmark & \checkmark & \checkmark & - &  2680.0 &   ~0.3716 \\
				\hline
			\end{tabular}
		}
	\end{table}

	\begin{table}
		\centering
		\caption{
			One-sided Wilcoxon signed-rank test for the NLST dataset.
			We compare the mean absolute error (MAE) of our best performing method (PseudoLabels) against the rest. \textsuperscript{*} indicates that the best performing method is better assuming a significance level of 5\%.
		}
		\label{tab:test_nlst}
		\scalebox{1.00}{
			\begin{tabular}{c|c|c|c|c|l}
				\hline
				~$\ell_{pen}$~ & ~$\ell_{ca\_rnk}$~ & ~Pseudo-labels~ & ~$\ell_{elr}$~  & ~Wilcox~ &  ~p-value~ \\
				\hline
				\checkmark & - & - & - &  4651.0 &  ~0.0023 \textsuperscript{*}\\
				\checkmark & \checkmark & - & - &  5252.0 &  ~0.0284 \textsuperscript{*}\\
				\checkmark & - & \checkmark & \checkmark &  2361.0 &  ~0.0099 \textsuperscript{*}\\
				\checkmark & \checkmark & \checkmark & - &  4807.0 &  ~0.2051 \\
				\checkmark & \checkmark & \checkmark & \checkmark &  4061.0 &  ~0.0593 \\
				\hline
			\end{tabular}
		}
	\end{table}

\end{document}


\title{Censor-aware Semi-supervised Learning for Survival Time Prediction from Medical Images}

\author{Paper 894}
\authorrunning{Paper 894}
\institute{Anonymous}

\maketitle


\begin{algorithm}
\caption{Pseudocode for the training procedure}
\begin{algorithmic}
\REQUIRE $\mathcal{D} = \{(\mathbf{x}_i,\mathbf{y}_i,\mathbf{e}_i\}_{i=1}^{|\mathcal{D}|} = \text{training data}$
\REQUIRE $h(\mathbf{x};\theta) = \text{survival model with parameters } \theta$
\REQUIRE $\alpha = \text{ELR loss strength}$
\REQUIRE $\beta = \text{Bin penalization strength}$
\REQUIRE $\delta = \text{Rank loss strength}$
\FOR{$k \text{ in } [1,k_{total}]$}
    \STATE $\textit{// Ratio of pseudo-labels from censored samples}$
    \STATE $m \leftarrow (1 + \cos{\pi} (1-k/k_{total})) / 2$
    \STATE $\textit{// Set of censored samples to use pseudo-labels}$
    \STATE $P \leftarrow random\_sample(\mathcal{D},m)$
    \FOR{$i \text{ in } P$} 
        \STATE $\textit{// Re-label censored sample}$
        \STATE $\tilde{\mathbf{t}}_i \leftarrow h(\mathbf{x}_i;\theta)$
        \STATE $\mathbf{y}_i \leftarrow \max(\mathbf{y}_i, \tilde{\mathbf{t}}_i)$
        \STATE $\mathbf{e}_i \leftarrow 1$
    \ENDFOR
    \STATE $\textit{// Update network parameters}$
    \FOR{\text{each minibatch} B}
        \FOR{$j \text{ in } B$}
            \STATE $\tilde{\mathbf{t}}_j \leftarrow h(\mathbf{x}_j;\theta)$
        \ENDFOR
        \STATE $loss \leftarrow \frac{1}{|\mathcal{D}|} \sum_{i=1}^{|\mathcal{D}|} \ell_{ca\_mse}(\tilde{\mathbf{t}}_i,\mathbf{y}_i,\mathbf{e}_i)$ \\
        \STATE \hspace{30pt} $+\alpha \cdot \ell_{elr}(\mathbf{p}_i) + \beta \cdot \ell_{pen}(\mathbf{p}_i)$
        \STATE \hspace{30pt} $+\frac{1}{|\mathcal{D}|}\sum_{j=1}^{|\mathcal{D}|}\delta \cdot \ell_{ca\_rnk}(\tilde{\mathbf{t}}_i,\tilde{\mathbf{t}}_j,\mathbf{y}_i,\mathbf{y}_j,\mathbf{e}_i,\mathbf{e}_j)$
        \STATE $\text{update } \theta \text{ using SGD}$ with $loss$
    \ENDFOR
\ENDFOR
\RETURN $\theta$
\end{algorithmic}
\label{alg:train}
\end{algorithm}


\begin{figure}[!t]
\centerline{\includegraphics[width=0.9\linewidth]{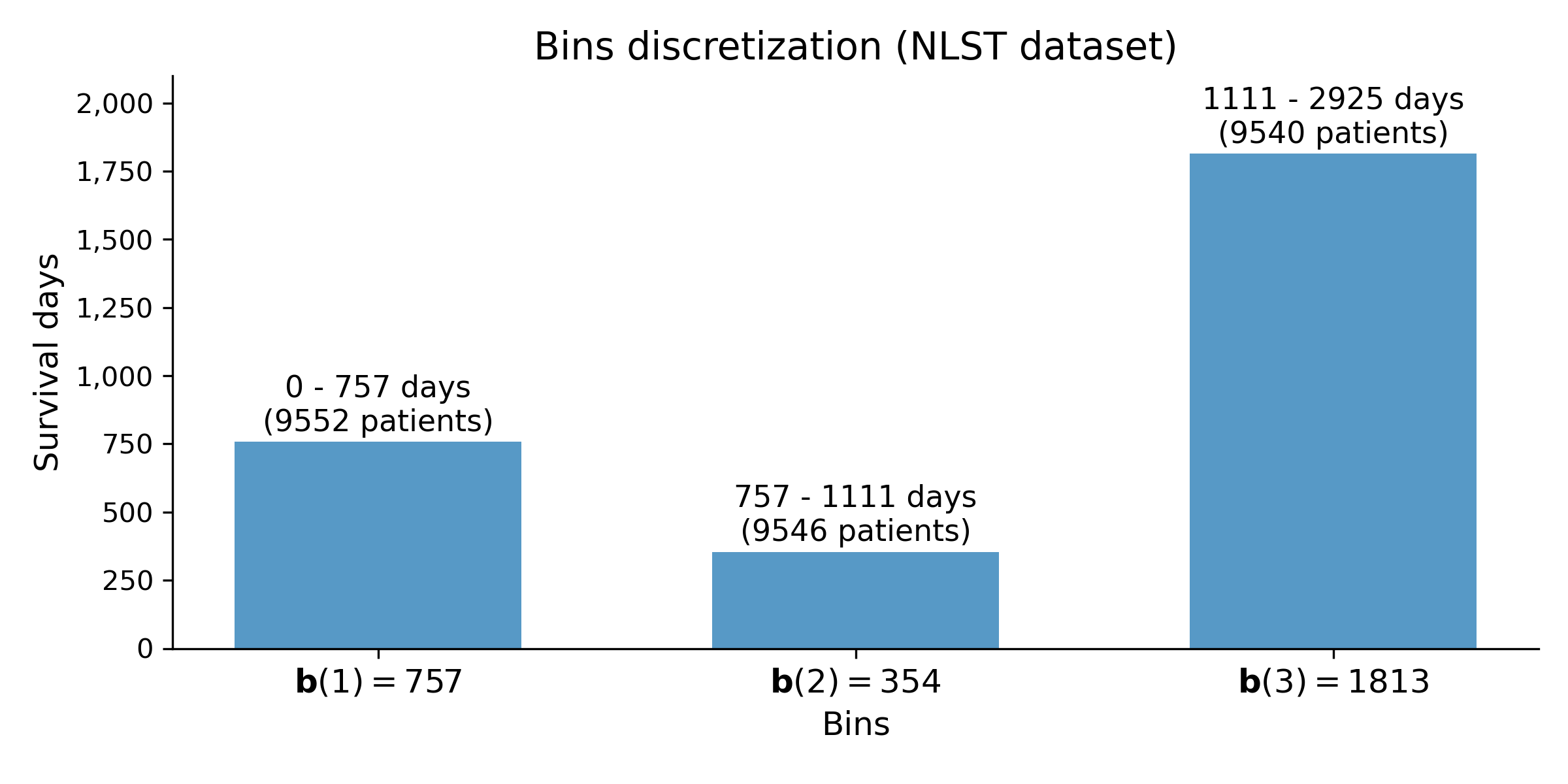}}
\caption{
We discretize the NLST dataset survival time into three bins, each representing a different amount of time, as depicted in the plot (e.g., $\mathbf{b}(1)=757$ days, $\mathbf{b}(2)=354$ days, and $\mathbf{b}(3)=1813$ days).
The discretization strategy aims to balance the number of patients per bin.
Note that this discretization strategy uses only the information in the training set.
}
\label{fig:bins}
\end{figure}

\begin{table}
\centering
\caption{
One-sided Wilcoxon signed-rank test for the TCGA-GM dataset.
We compare the mean absolute error (MAE) of our best performing method ($\text{Pseudolabels} + \ell_{ca\_rnk} + \ell_{elr}$) against the rest. \textsuperscript{*} indicates that the best performing method is better assuming a significance level of 5\%.
}
\label{tab:test_tcga}
\scalebox{1.00}{
\begin{tabular}{c|c|c|c|c|l}
\hline
~$\ell_{pen}$~ & ~$\ell_{ca\_rnk}$~ & ~Pseudo-labels~ & ~$\ell_{elr}$~  & Wilcox &  p-value \\
\hline
\checkmark & - & - & - &  2157.0 &   0.0115 \textsuperscript{*} \\
\checkmark & \checkmark & - & - &  2069.0 &   0.0113 \textsuperscript{*} \\
\checkmark & - & \checkmark & - &  2383.0 &   0.0579 \\
\checkmark & - & \checkmark & \checkmark &  2633.0 &   0.3164 \\
\checkmark & \checkmark & \checkmark & - &  2680.0 &   0.3716 \\
\hline
\end{tabular}
}
\end{table}

\begin{table}
\centering
\caption{
One-sided Wilcoxon signed-rank test for the NLST dataset.
We compare the mean absolute error (MAE) of our best performing method (PseudoLabels) against the rest. \textsuperscript{*} indicates that the best performing method is better assuming a significance level of 5\%.
}
\label{tab:test_nlst}
\scalebox{1.00}{
\begin{tabular}{c|c|c|c|c|l}
\hline
~$\ell_{pen}$~ & ~$\ell_{ca\_rnk}$~ & ~Pseudo-labels~ & ~$\ell_{elr}$~  & Wilcox &  p-value \\
\hline
\checkmark & - & - & - &  4651.0 &  0.0023 \textsuperscript{*}\\
\checkmark & \checkmark & - & - &  5252.0 &  0.0284 \textsuperscript{*}\\
\checkmark & - & \checkmark & \checkmark &  2361.0 &  0.0099 \textsuperscript{*}\\
\checkmark & \checkmark & \checkmark & - &  4807.0 &  0.2051 \\
\checkmark & \checkmark & \checkmark & \checkmark &  4061.0 &  0.0593 \\
\hline
\end{tabular}
}
\end{table}
